\documentclass[letterpaper]{article} 
\usepackage{aaai2026}  
\usepackage{times}  
\usepackage{helvet}  
\usepackage{courier}  
\usepackage[hyphens]{url}  
\usepackage{graphicx} 
\usepackage{natbib}  
\usepackage{caption} 
\usepackage{algorithm}
\usepackage{algorithmic}

\usepackage{newfloat}
\usepackage{listings}
\DeclareCaptionStyle{ruled}{labelfont=normalfont,labelsep=colon,strut=off} 
\floatstyle{ruled}
\newfloat{listing}{tb}{lst}{}
\floatname{listing}{Listing}
\usepackage{amsmath}

\nocopyright

\title{MAGIC: Multi-Agent Argumentation and Grammar
Integrated Critiquer}
\author{
    Joaqu\'in Jord\'an\equalcontrib,
    Xavier Yin\equalcontrib,
    Melissa Fabros\equalcontrib,
    Gireeja Ranade,
    Narges Norouzi
}
\affiliations{
    UC Berkeley\\

    \{jjordanoc, nzxyin, melissa.fabros, gireeja, norouzi\}@berkeley.edu
}

\begin{document}

\maketitle 

\let\thefootnote\relax\footnotetext{Preprint. Accepted into EAAI 2026.}

\begin{abstract}
Automated Essay Scoring (AES) and Automatic Essay Feedback (AEF) systems aim to reduce the workload of human raters in educational assessment. However, most existing systems prioritize numerical scoring accuracy over feedback quality and are primarily evaluated on pre-secondary school level writing. This paper presents Multi-Agent Argumentation and Grammar Integrated Critiquer (MAGIC), a framework using five specialized agents to evaluate prompt adherence, persuasiveness, organization, vocabulary, and grammar for both holistic scoring and detailed feedback generation. To support evaluation at the college level, we collated a dataset of Graduate Record Examination (GRE) practice essays with expert-evaluated scores and feedback. MAGIC achieves substantial to near-perfect scoring agreement with humans on the GRE data, outperforming baseline LLM models while providing enhanced interpretability through its multi-agent approach. We also compare MAGIC's feedback generation capabilities against ground truth human feedback and baseline models, finding that MAGIC achieves strong feedback quality and naturalness.
\end{abstract}

\begin{links}
     \link{Supplement}{https://github.com/magic-aes/MAGIC}
\end{links}

\begin{figure*}[ht!]
  \centering
    \includegraphics[width=0.85\linewidth, keepaspectratio]{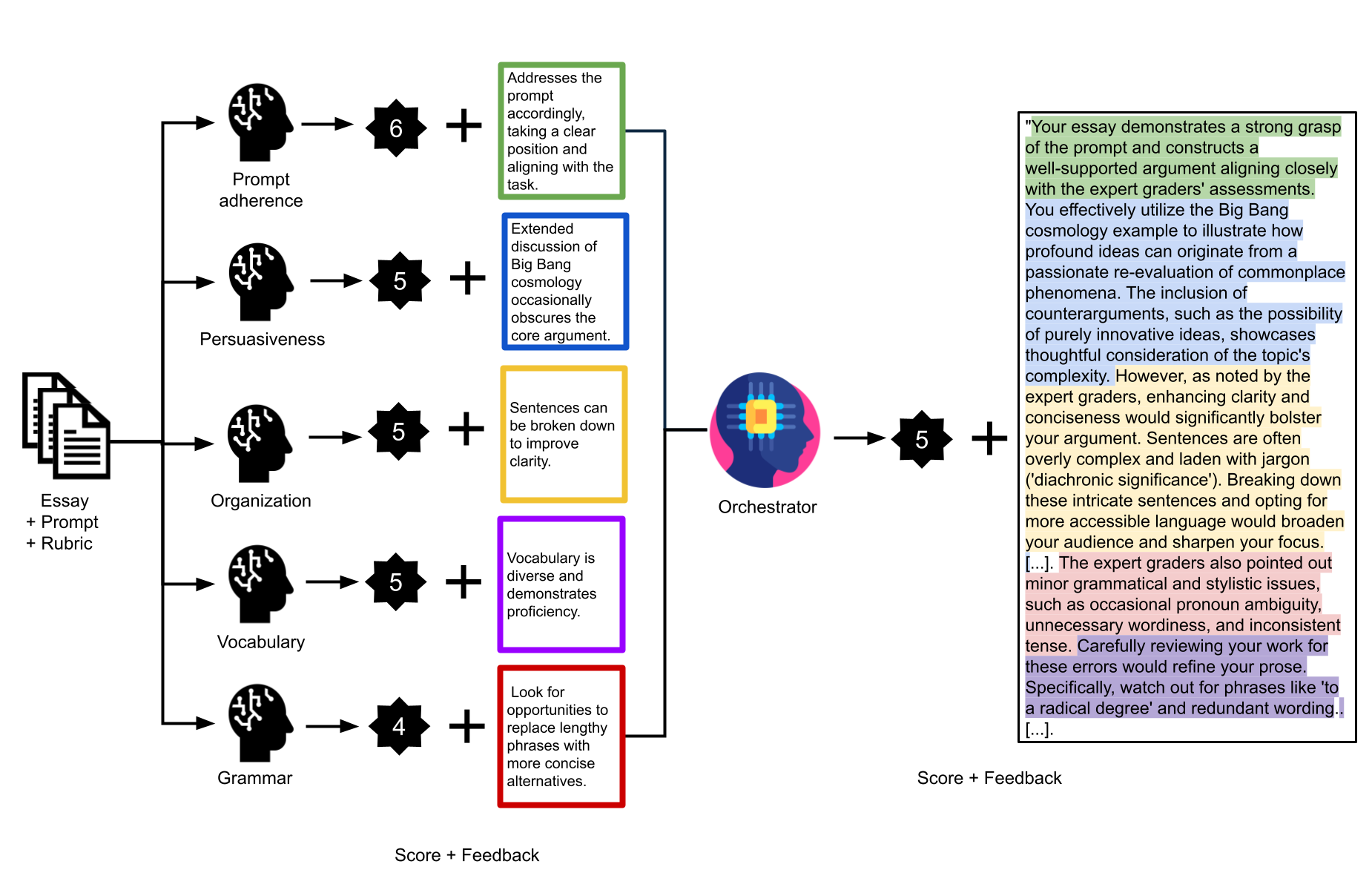}
    \caption{\textbf{MAGIC AES Feedback and Scoring Pipeline.} Each agent (prompt adherence, persuasiveness, organization, vocabulary, and grammar) scores the essay separately and provides feedback for their assigned trait. The orchestrator merges the agents' results into a holistic score and combined feedback.}
    \label{fig:magic-aes}
\end{figure*}

\section{Introduction}

Automated Essay Scoring (AES) and Automated Essay Feedback (AEF) are gaining importance in educational assessment, aiming to replicate human evaluation of written work based on content, coherence, grammar, and style \cite{Dikli_2006}. While AES systems have achieved notable success in predicting human‐assigned numerical scores, generating meaningful, personalized essay feedback at scale remains an open problem \cite{behzad2024assessing}. 

The limitations of commonly used datasets for multi-trait scoring and feedback evaluation have created a gap in AES and AEF research. The NLP research community has invested in compiling essay datasets such as TOEFL11~\cite{blanchard-et-al-toefl11}, ASAP++~\cite{mathias-bhattacharyya-2018-asap}, and ICLE++~\cite{li-ng-2024-icle}. However, most of these datasets do not provide ground truth per-trait scores, feedback, or exist behind a paywall. Furthermore, many of the commonly used datasets are collected from writers as part of English as a second language (L2) exams or come from high-school populations of native speakers of English (L1), such as ASAP~\cite{hewlett2012asap} and PERSUADE 2.0~\cite{CROSSLEY2024persuade2}. Feedback to L2 English learners will focus on different qualities than learners with native English who are still developing their writing and critical thinking skills \cite{PAN201660}. As for essays written by L1 speakers, the reliance on ASAP-based evaluation limits robust exploration of how to improve AES and feedback to L1 English writers above the 10th grade level \cite{li-ng-2024-icle}. 

AES systems are currently most useful in large-scale testing environments where thousands of writing samples must be scored quickly. In classrooms, where students are learning to produce effective essays, numerical scores or grades alone do not improve learning \cite{guskey_grades_2019}. Writing and argument are central to both educational development and intellectual growth. \citet{Riddell_2015} therefore advocates for frequent feedback with increased writing opportunities as a recipe for higher learning outcomes. However, scaling instructors’ feedback capacity without sacrificing quality remains an ongoing challenge \cite{page-grading}. Since our educational aims should prioritize teaching ``intelligent humans'' over intelligent tutoring systems \cite{Baker2016}, AI feedback on writing tasks must be integrated thoughtfully and carefully to enhance rather than diminish learners' cognitive engagement.

When using generative AI for feedback generation, like \citet{favero_leveraging_2025}, we found frontier large language  models such as OpenAI's ChatGPT and Anthropic's Claude capable of discerning nuances of argument and grammar. However, these enterprise-level API-based models are expensive for school systems to support and difficult to guarantee privacy and accessibility. Therefore, smaller open-sourced models are advantageous for their computational efficiency and open-weights. Educators and administrators can deploy these systems locally to ensure student privacy, model observability, and prediction explainability.

We introduce Multi-Agent Argumentation and Grammar Integrated Critiquer (MAGIC), a generic framework for zero-shot multi-trait AES using independent small LLM agents to grade and provide feedback for each writing dimension of a rubric. 

Our experimentation yielded the following insights and contributions to AES and AEF:
\begin{enumerate}
    \item We propose the use of a compiled Graduate Record Examination (GRE) dataset for assessing AES and AEF model performance in college-level argumentative writing, exhibiting high-quality argumentation, and including feedback ground truths.
    \item We present MAGIC, a zero-shot multi-agent framework for AES and AEF, generalizing to various argumentative essay prompts, responses, and rubrics without the need for fine-tuning or training. MAGIC outperformed our single agent baseline in scoring, showing improvements on both small- and medium-sized open-weight LLMs.
    \item We provide a systematic analysis of MAGIC’s feedback using various metrics to help assess feedback specificity and relevance with the source essay. 
\end{enumerate}

All complete prompts and rubrics mentioned in this work, as well as dataset sources, are available in our online supplement.

\section{Prior Work}

The fields of Automated Essay Scoring (AES) and Automated Essay Feedback (AEF) have evolved from statistical modeling approaches \cite{page-stats-grading} to modern LLM-based systems, revealing persistent challenges in balancing scoring accuracy with meaningful feedback generation.

\subsection{Early Approaches}
Earlier NLP studies employed Naive Bayes, Support Vector Machines, and Decision Trees for scoring, with Latent Semantic Analysis for feedback generation \cite{Liu_AES_feedback_2017}. While handcrafted features offered some interpretability, they remain costly to create and generalize poorly across different prompts and essay types \cite{misgna2025survey}. The transition to deep learning marked significant improvements, with transformer-based approaches like $R^2BERT$ outperforming most traditional models \cite{yang2020enhancing}.

\subsection{Datasets}
The ASAP corpus \cite{hewlett2012asap} became the de facto AES benchmark, containing essays from US students in grades 7-10, with extensions like ASAP++ \cite{mathias-bhattacharyya-2018-asap} adding multi-trait annotations. Other corpora covering different demographics, TOEFL11 \cite{blanchard-et-al-toefl11}, CLC-FCE \cite{yannakoudakis-etal-2011-new}, and ICLE++ \cite{li-ng-2024-icle}, remain underutilized. Critically, research in feedback generation remains limited compared to AES \cite{misgna2025survey, wang2022automated}, with existing datasets focusing primarily on lower-grade or second-language writing, creating a significant gap for college-level argumentative writing evaluation.

\subsection{Large Language Models and Zero-shot AES}
LLMs initially showed weak correlation with human evaluations on the ASAP dataset, though prompt engineering with few-shot examples improved alignment \cite{kundu2024largelanguagemodelsgood}. \citet{naismith-etal-2023-automated} achieved QWK nearing 0.80 with GPT-4 using task instructions and rubrics, while \citet{seßler2024aigradeessayscomparative} found OpenAI o1 outperformed other LLMs across multiple traits. However, GPT-4 performance still does not surpass modern AWE methods \cite{yancey2023rating}.

Recent zero-shot approaches address accessibility limitations of few-shot strategies. \citet{lee_unleashing_2024} introduced Multi Trait Specialization (MTS), decomposing writing proficiency into distinct traits with trait-specific evaluation, achieving QWK gains of 0.437 on TOEFL11. \citet{shibata_lces_2025} proposed comparative scoring using pairwise judgments to address bias issues in direct scoring methods.

\subsection{Feedback Generation Challenges}
Early feedback work achieved only 50\% appropriate response rates with seq2seq models \cite{hanawa2021exploring}. \citet{Villalon_Glosser_2008} developed Glosser, an LSA-based model for topic-specific feedback, though it exhibited bias toward longer sentences despite lacking coherence. Recent studies found GPT-4 error-prone in Grammar Error Explanation with frequent hallucinations \cite{song2023gee}. Joint scoring and essay feedback generation using Chain-of-Thought achieved only 0.533 QWK \cite{stahl2024exploring}, far below state-of-the-art scoring baselines of 0.79 \cite{yang2020enhancing}. While some studies show promise for EFL learners \cite{han2024llm} and comparisons with forum feedback \cite{behzad2024assessing}, fundamental challenges remain in providing interpretable, high-quality feedback that matches human expertise.

This background motivates the need for transparent, multi-agent frameworks that provide both accurate scoring and interpretable feedback.

\section{Dataset}

We collated exam preparation material published for free by Educational Testing Services (ETS) for Graduate Record Examination (GRE) self-study and exam transparency. The GRE exam consists of multiple choice questions and an essay response, and test-takers are most often students writing at the post-secondary or university level who have more experience and facility with the argumentative essay genre, consisting of a mix of both L1 and L2 English writers. 

This ground truth evaluation set is comprised of ``Analytical Writing Sample Essays with Commentaries'' from 48 GRE essays in eight essay prompts, each with holistic scores between 1--6 and associated human qualitative feedback based on a provided rubric \cite{ETS2023GRETest1WritingResponses}.

Furthermore, in order to have corresponding ground truth not only for holistic scores but also on the trait level, we created five trait-based sub-rubrics (T) based on the holistic rubric, where each feature can be scored between 1--6:
\begin{itemize}
    \item[] T1. Quality of the response to the prompt \\instructions
    \item[] T2. Considering the complexities of the issue
    \item[] T3. Organizing, developing, and expressing ideas
    \item[] T4. Vocabulary and sentence variety
    \item[] T5. Grammar and mechanics 
\end{itemize}

One of the authors with previous experience in GRE essay grading annotated each essay with a 1--6 score for every trait. 

We argue that this dataset is more representative of the typical workloads teachers might encounter in the higher-education classroom, being unable to train large neural AES models because of the scarce data and having to rely on zero-shot approaches.

\section{Methodology}
\label{ch:methodology}

\subsection{MAGIC: A Multi-Agent Approach}
MAGIC (Multi-Agent Argumentation and Grammar Integrated Critiquer) is an adaptable framework for zero-shot multi-trait AES. MAGIC decomposes holistic assessment into specialized LLM agents, each evaluating a distinct writing dimension specified by a rubric before an orchestrator agent produces the final score and feedback. Our proposed framework adapts to different essay types and rubrics by having modular prompts for each agent.

Unlike previous work using LLMs for holistic scoring through single prompting strategies, our approach isolates each rubric trait in a separate agent interaction. This method, similar to \cite{lee_unleashing_2024}, enables deeper model reasoning for individual traits. We improve upon this premise by introducing an orchestrator agent that integrates trait-level assessment into holistic scores and feedback, improving scoring agreement with human raters. Our summarized approach is shown in Figure~\ref{fig:magic-aes}.

For this work, we instantiate MAGIC with the GRE argumentative writing evaluation task, as described in the Dataset section. We employ five specialized agents corresponding to our rubric traits T1--T5. Agents T1--T3 evaluate the argumentative qualities of the essay, agent T4 assesses vocabulary, and agent T5 focuses on grammar. The orchestrator then receives these assessments and synthesizes the final feedback and score. All agents are aware of the given essay's content and prompt, but each specialized agent only has access to their trait's rubric to ensure isolation.

While our evaluation focuses on argumentative essays,  our framework can be extended to other types of essays (e.g.~narrative essays) by replacing the model's prompts and providing new rubrics for the desired task.

\subsection{Evaluation}
\label{sec:evaluation}
We built MAGIC with small open-source instruction-tuned LLMs in mind, given their accessibility in educational environments and competitive performance in AEF-related tasks \cite{favero_leveraging_2025}. The LLMs used for our evaluation were: Llama 3.1 8B, Gemma 3 12B, and Gemma 3 27B. Our baseline model consists of a single LLM with a CoT prompt producing a holistic score and feedback. 

Aligning with standard AES research, we apply Quadratic Weighted Kappa (QWK) to measure score agreement, which penalizes larger score differences more heavily than smaller ones (Table~\ref{tab:qwk-scores}).

\begin{table}[ht!]
    \centering
    \begin{tabular}{cc}
    \hline
        \textbf{QWK score} & \textbf{Agreement} \\
    \hline
        $\leq$ 0 & None\\
        0.01 -- 0.20 & Slight\\
        0.21 -- 0.40 & Fair\\
        0.41 -- 0.60 & Moderate\\
        0.61 -- 0.80 & Substantial\\
        0.81 -- 0.99 & Near-Perfect\\
        1.00 & Perfect\\
    \hline
    \end{tabular}
    \caption{\textbf{Explanation of QWK score ranges.} The interpretation of different QWK score ranges used in our evaluation, as presented in \citet{landis1977measurement}.}
    \label{tab:qwk-scores}
\end{table}

To demonstrate its capabilities, we evaluated MAGIC against our baseline results against the collated GRE essay set for both holistic scoring and feedback generation. We report QWK and RMSE results for both baseline and MAGIC variants. Beyond holistic scoring, we also measure per-trait QWK of the independent agents in MAGIC against human-annotated per-trait ground truth scores. 

Moreover, to assess the quality of the feedback generated by our models, we annotated human--LLM feedback pairs. We used the following evaluation criteria (C) to assess feedback quality~\cite{behzad2024assessing}:

\begin{itemize}
\label{feedback-criteria}
    \item[] C1. Which is more relevant to the essay content?
    \item[] C2. Which is better at highlighting weakness?
    \item[] C3. Which is better at highlighting strengths?
    \item[] C4. Which is more specific and actionable?
    \item[] C5. Which is more helpful for a student overall?
\end{itemize}

Additionally, we assessed MAGIC's feedback in the absence of human annotators using an LLM judge \cite{zheng2023judging}, and we obtained aligned feedback judgments. For this judge agent, we used OpenAI o4-mini reasoning model, via OpenAI's API, with the ``medium'' reasoning level for all experiments.  

\subsection{Experiment Infrastructure}
We used one NVIDIA A100 GPU with 80GB of VRAM, for 10 hours to run the scoring and feedback generation experiments. We used vLLM for GPU-efficient LLM inference.

\section{Results}
\begin{table*}[t!]
\centering
\begin{tabular}{l|rrr|rrr}
\hline
\textbf{Model} & \textbf{QWK} base $\uparrow$ & \textbf{QWK} MAGIC $\uparrow$ & $\Delta$\textbf{QWK} $\uparrow$ & \textbf{RMSE} base $\downarrow$ & \textbf{RMSE} MAGIC $\downarrow$ & $\Delta\textbf{RMSE}$ $\downarrow$ \\ \hline
gemma3-12b-it & 0.607 & 0.762 & \textbf{0.155} & 1.263 & \textbf{0.994} & $-$\textbf{0.269}\\
gemma3-27b-it & \textbf{0.613} & \textbf{0.766} & 0.153 & \textbf{1.250} & 1.010 & $-$0.240\\
llama3.1-8b-it & 0.605 & 0.731 & 0.126 & 1.307 & 1.163 & $-$0.144\\ \hline
\end{tabular}
\caption{\textbf{AES performance comparison.} Quadratic Weighted Kappa (QWK) and Root Mean Squared Error (RMSE) are measured against the ground truth GRE scores. The QWK and RMSE results for both baseline and MAGIC as well as the change in each metric ($\Delta$) are shown for the three chosen LLMs. The best result for each column has been bolded.}
\label{tab:gre-aes-feedback}
\end{table*}

\subsection{Scoring Agreement Against Humans}

We compared QWK scores and feedback quality across all our tested LLM models for baseline and MAGIC. Our results in Table~\ref{tab:gre-aes-feedback} show all three of the models saw an increase in QWK between baseline and MAGIC configurations. The largest increase was observed in Gemma 3 12B, from moderate (0.607) to substantial (0.762). Gemma 3 27B had the highest QWK for both base and MAGIC configurations, 0.613 and 0.766 respectively. As for RMSE, Gemma 3 12B had the largest improvement from baseline to MAGIC, decreasing RMSE by 0.269. Gemma 3 12B MAGIC outperforms the Gemma 3 27B MAGIC marginally on RMSE despite the 27B model having better baseline RMSE. These results highlight MAGIC's effectiveness for AES tasks.

\begin{figure}[h]
    \centering
    \includegraphics[width=0.45\textwidth]{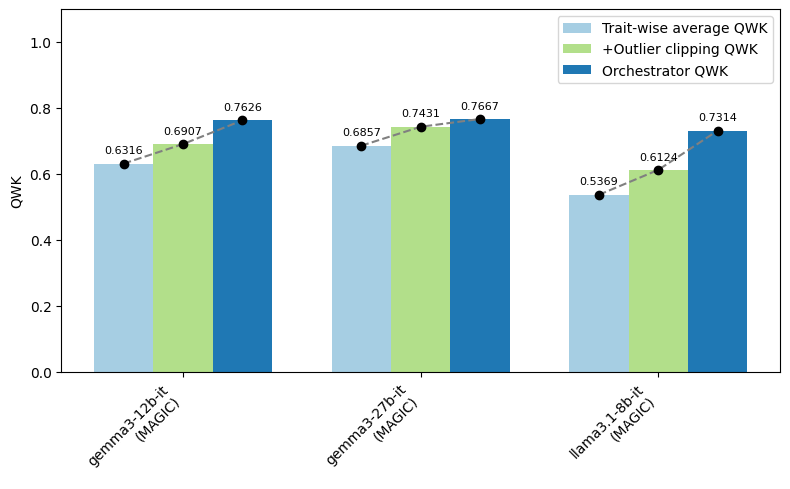}
    \caption{\textbf{Comparison of holistic score QWK} between taking the average across traits (Trait-wise QWK), adding an additional outlier clipping and scaling stage (+Outlier clipping QWK), as shown in \citet{lee_unleashing_2024}, and using an orchestrator agent (Orchestrator QWK).}
    \label{fig:avg-vs-orchestrated}
\end{figure}

We designed an orchestrator agent to produce a holistic score, as opposed to using a simpler aggregation scheme on the per-trait scores. We compared its score against human scores across different aggregation strategies: simple averaging over the trait scores, average with outlier clipping and scaling as shown in \citet{lee_unleashing_2024}. We observed that the holistic score readout provided by these other strategies yields lower concordance with human scores. Instead, having the orchestrator consider all the agents' scores and feedback and then produce its own holistic score shows improved concordance, as shown in Figure \ref{fig:avg-vs-orchestrated}. This might suggest unequal weighting of the traits by the human graders, which the orchestrator can more faithfully capture.

\begin{figure}[ht!]
    \centering
    \includegraphics[width=0.48\textwidth]{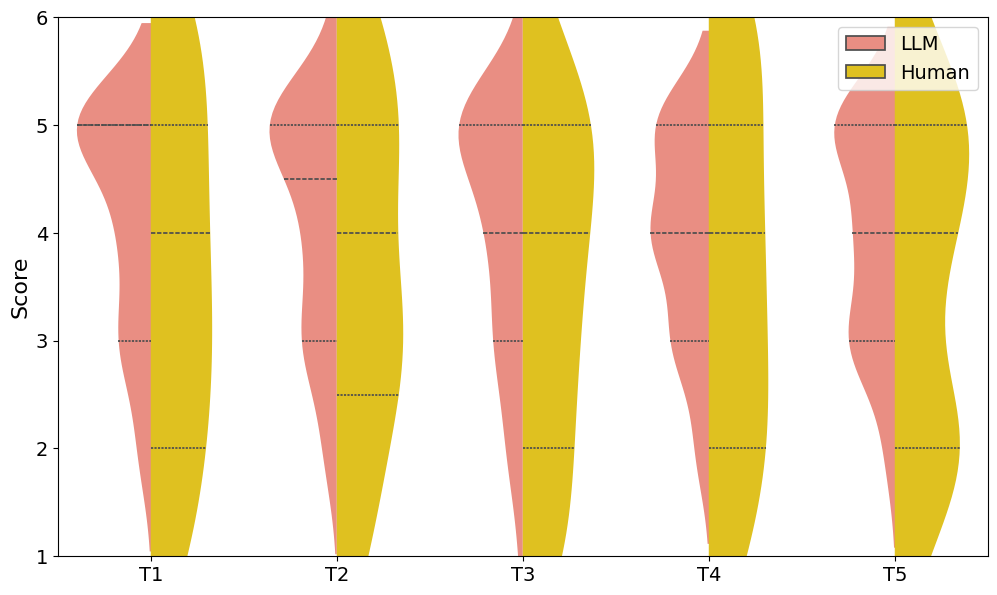}
    \caption{\textbf{Per-trait score distributions} for Gemma 3 27B LLM and human annotated ground-truth on the GRE dataset. Quartiles highlighted as dotted lines.}
    \label{fig:domain-scores}
\end{figure}

Moreover, on the agent scoring level, we see that the agents' scores tend to avoid extremes but roughly follow the human score distribution (Figure~\ref{fig:domain-scores}). We also see some of the grade inflation tendencies associated with LLM-based AES systems. Our experiments with generating feedback and scores holistically and on a feature-trait level corroborate recent findings where LLM graders were more likely to assign scores near the average, and humans assigned a wider range of scores \cite{the_learning_agency_2025}. 

\begin{figure}[ht!]
    \centering
    \includegraphics[width=0.42\textwidth]{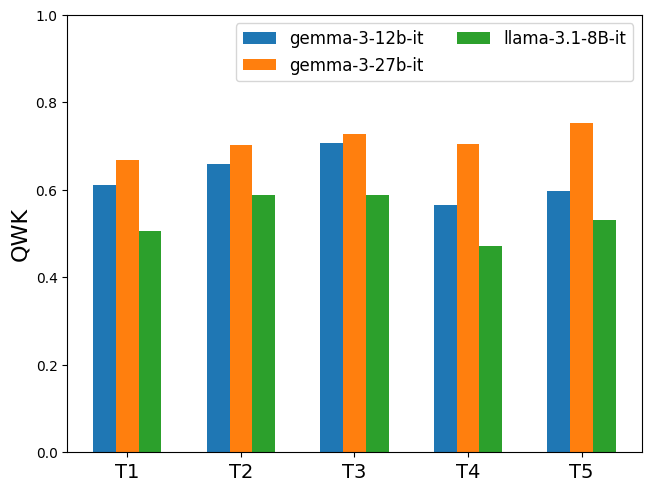}
    \caption{\textbf{Per-trait QWK of MAGIC independent agents across different base LLMs.}  Our writing dimension traits (T1--T5) are as described in our  methodology.}
    \label{fig:gre-trait-human}
\end{figure}

Further breaking down the QWK score, Figure~\ref{fig:gre-trait-human} shows the per-trait QWK between LLMs and human ground truth. We evaluated the per-trait scoring capabilities of each of our agents against per-trait human ground truths, reaching moderate to substantial agreement between MAGIC scores and human scores on each of the traits. We notice an LLM trend to score higher on argumentative-related criteria (T1--T3) than on vocabulary (T4) and grammar (T5). Moreover, Gemma 3 27B offers the highest agreement with ground truth scores, consistent with our results on holistic score QWK.

%
%

\subsection{Comparison Against Human Feedback}

For subsequent experiments in AEF, we selected Gemma 3 27B baseline and Gemma 3 27B with MAGIC as our LLM models.

\begin{figure}[ht!]
    \centering
    \includegraphics[width=0.48\textwidth]{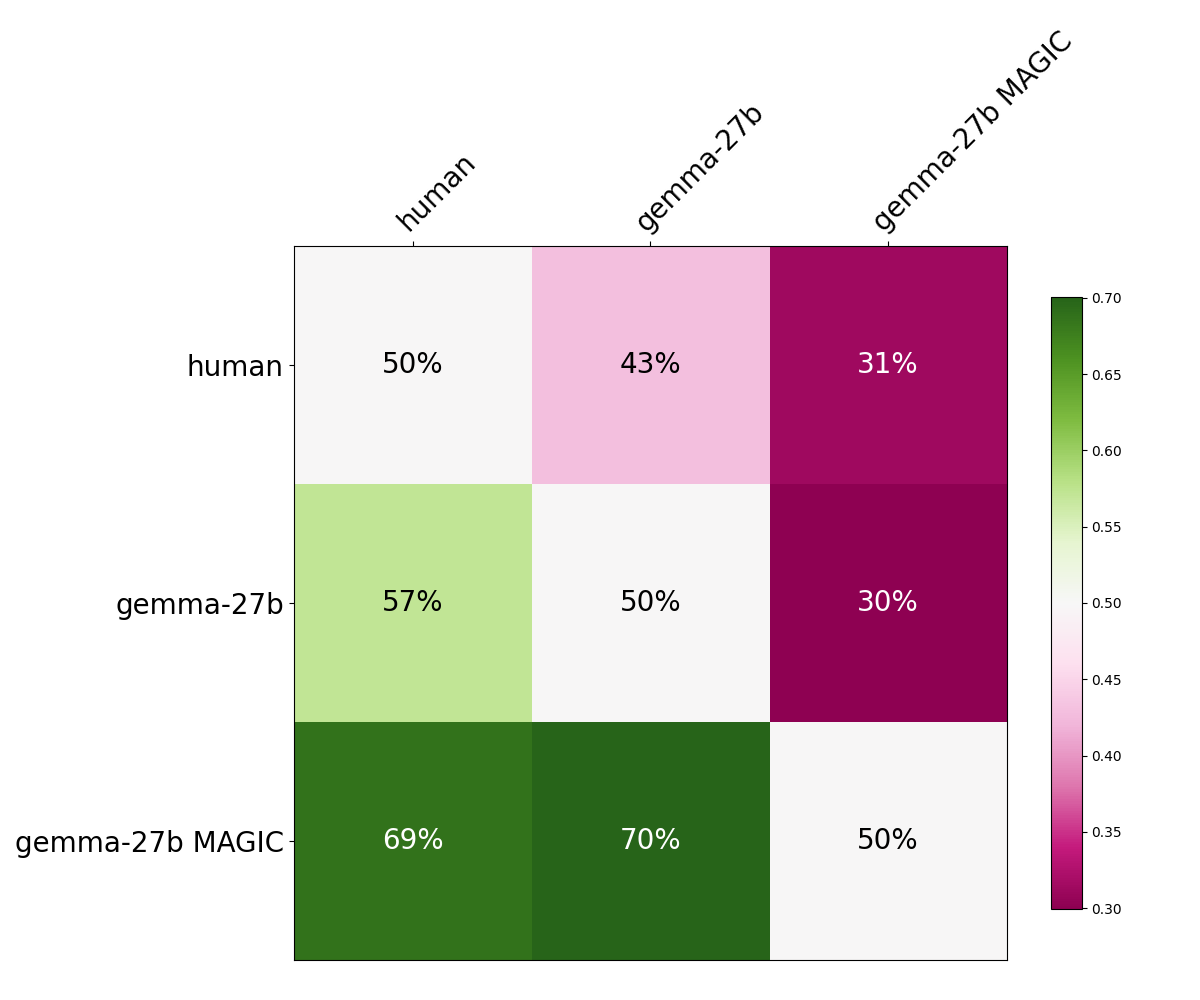}
    \caption{\textbf{Head-to-head Model Feedback Win-rates as Rated by a Judge LLM (o4-mini).} Value at row $i$ and column $j$ denotes the average win-rate of row $i$ over column $j$ across all 5 criteria (C1--C5).}
    \label{fig:feedback-arena-matrix}
\end{figure}

To compare the quality of the generated feedback, we paired models in an A--B test ``battle,'' similar to Chatbot Arena \cite{chiang2024chatbot}, using an LLM judge following the previous feedback assessment criteria (C1--C5) as explained in the Methodology Section. We compute the average majority win-rate over all 5 criteria. Results are shown in Figure \ref{fig:feedback-arena-matrix}. Due to previous works such as \cite{zheng2023judging} demonstrating that strong LLMs are capable of having a high level of agreement with humans, we believe that this judging strategy provides a scalable way to evaluate essay feedback. At the same time, the authors note that strong LLMs tend to also prefer LLM-generated responses \cite{zheng2023judging}. Further analysis of the Judge model is in the ``Judging Feedback'' section.

Gemma 3 27B without MAGIC wins marginally more often than losing with a 57\% winrate against the ground-truth human-written GRE feedback. With MAGIC, the winrate jumps to 69\% against human feedback and 70\% against baseline Gemma 3 27B. However, LLM-judges have been shown to have biases towards positive evaluations \cite{koutcheme_evaluating_2024}, so we further inspect and analyze the outputs of our best performing model: Gemma 3 27B with MAGIC.

\subsection{Feedback Characteristics}

We observe that MAGIC feedback was statistically on par with the human feedback on the word count axis. MAGIC's feedback length saw only 14\% difference. Human feedback was on average longer, however, MAGIC feedback remained more consistent across score points. MAGIC averaged up to twice as many words than human evaluators for the lowest scoring samples.

While LLM-generated feedback demonstrates comparable lexical diversity (MATTR: 0.76-0.78 vs 0.71), human feedback exhibits a 60\% larger vocabulary (1,705 vs 1,050-1,127 unique words), higher entropy (8.52 vs 8.11-8.25), and greater use of unique expressions (.534 vs .49-.51 hapax legomena). This suggests that while LLMs maintain consistent linguistic variety within individual feedback instances, human experts draw from a richer and less predictable lexical repertoire across the corpus (Table \ref{tab:diversity}).

\begin{table}[ht!]
\centering
\begin{tabular}{l|l|l|l}
\hline
\textbf{Metric} & \textbf{Human} & \textbf{MAGIC} & \textbf{Baseline} \\
\hline
Vocabulary Size & 1,705 & 1,127 & 1,050 \\
MATTR & 0.7081 & 0.7764 & 0.7620 \\
Repetition Rate (\%) & 90.3 & 91.6 & 91.2 \\
Hapax Ratio (\%) & 53.4 & 49.2 & 51.0 \\
Entropy & 8.5220 & 8.2505 & 8.1075 \\
\hline
\end{tabular}
\caption{\textbf{Corpus wide Human-LLM lexical diversity} LLM feedback shows good text diversity but reveals lower overall vocabulary richness against human feedback}
\label{tab:diversity}
\end{table}

The individual agent feedback is centered on how to improve the specific essay and provides specific and actionable recommendations for this essay sample in a single turn alongside a numerical score. Peer research such as ~\citet{lee_unleashing_2024} required a second conversation turn to collect essay quotes to justify the trait score. MAGIC appears to avoid generic assessments by tailoring its feedback on how to improve the essay at hand along a specific trait. 
We also observe MAGIC producing feedback that spent more time highlighting the essay sample's strengths (C3) over the human feedback, which appeared to better highlight the essay's weaknesses (C2). It also tends to reference the other agents as expert graders or aspect evaluators as part of its rationale for its own commentary.

We measure MAGIC's ability to ``personalize'' its feedback by looking at Jaccard Similarity between the source essay and the feedback text. LLMs are known to provide generic non-specific advice, and we test MAGIC's feedback specificity by comparing its Jaccard scores against an unrelated and a random text. Jaccard Similarity measures direct overlap between words in the feedback and essay to which it is responding. We see this metric as a proxy for textual references and quotations that the feedback text uses to engage with the specificity of the essay. 

\begin{figure}[h!]
    \centering    \includegraphics[width=0.48\textwidth]{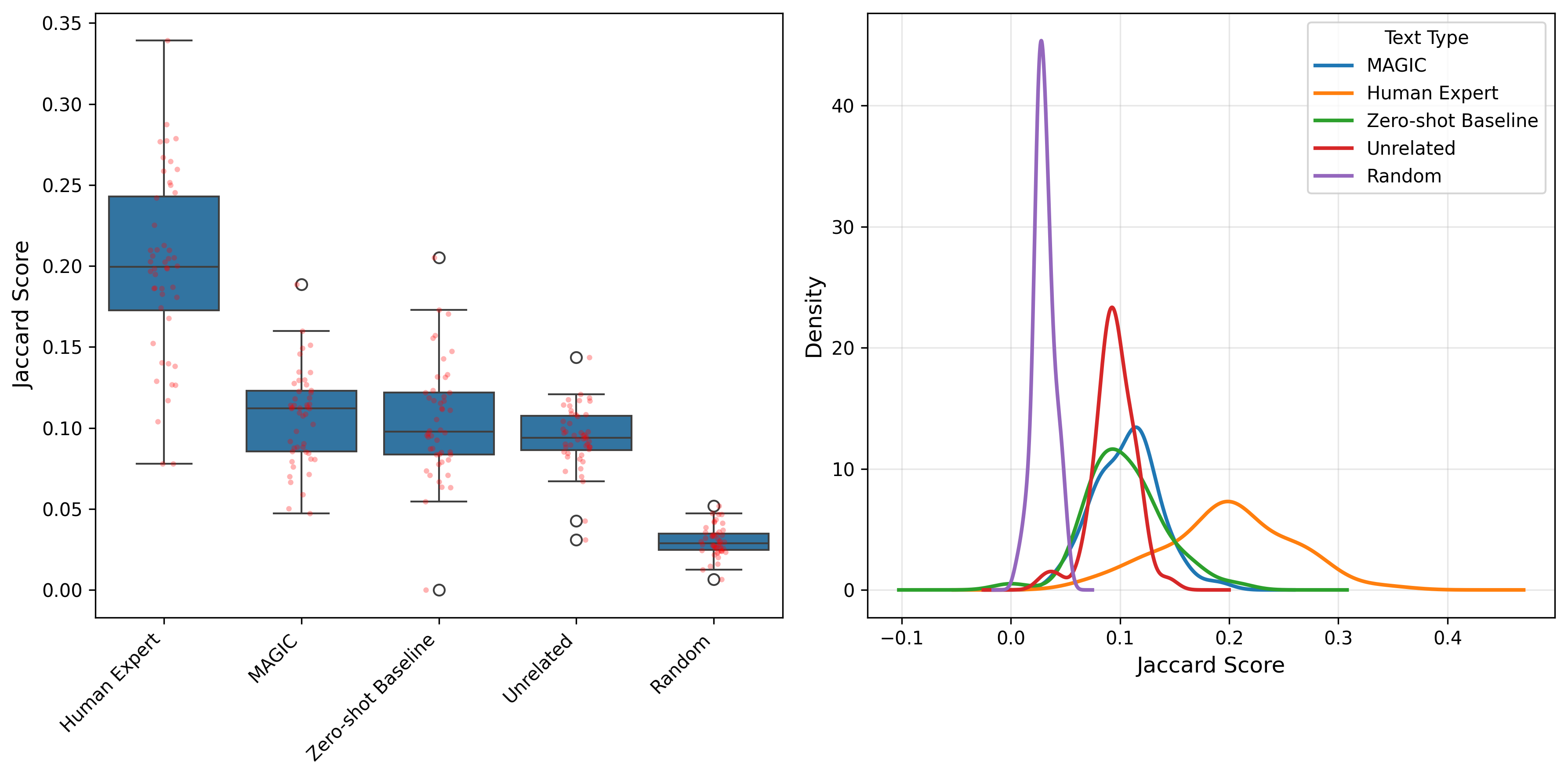}
    \caption{\textbf{Jaccard Similarity between different feedback conditions and essay text.} Left: Red dots represent essays, y-axis is Jaccard Similarity. Right: Distribution of Jaccard scores (KDE). Unrelated and random are baselines with an unrelated text passage and totally random text respectively.}
    \label{fig:jaccard-similarity}
\end{figure}

As seen in Figure \ref{fig:jaccard-similarity}, all types of feedback are clear improvements over random text. While the orchestrated feedback closes the gap toward the human level specificity, it is a statistically moderate improvement from the unrelated text. The baseline text offers a low statistical advantage over valid unrelated text.



We also evaluate alignment between LLM and human feedback using text similarity metrics (Table 4). Overall, LLM feedback aligns well with human feedback, as reflected in moderately high BERTScore F1 values—a context-aware metric believed to correlate with human judgments \cite{rehman2025evaluating}. Differences among MAGIC, baseline, or other LLM outputs are not significant, suggesting our prompts already elicit human-like assessments. Meanwhile, ROUGE-L, an LCS-based metric, remains low, indicating that LLM wording choices diverge perceptibly yet consistently, likely due to correct adherence to the shared prompt.

\begin{table}[ht!]
\begin{tabular}{l|l|l}
\hline
\textbf{Model}          & \textbf{ROUGE-L F1}  & \textbf{BERTScore F1} \\ \hline
gemma-27b      & \textbf{0.175 $\pm$ 0.007}               & 0.851 $\pm$ 0.003                 \\
gemma-27b MAGIC & 0.159 $\pm$ 0.005               & 0.844 $\pm$ 0.002                 \\ \hline
gemma-12b      & 0.174 $\pm$ 0.008               & \textbf{0.853 $\pm$ 0.002}                 \\
gemma-12b MAGIC & 0.162 $\pm$ 0.006               & 0.845 $\pm$ 0.002                 \\ \hline
llama-8b       & 0.163 $\pm$ 0.009               & 0.852 $\pm$ 0.003                 \\
llama-8b MAGIC  & 0.166 $\pm$ 0.018               & 0.849 $\pm$ 0.006                 \\ \hline
\end{tabular}
\caption{\textbf{Human-LLM feedback alignment} between LLM feedback outputs and ground truth feedback. Scores are reported as mean with 95\% confidence intervals. Best results per-column are highlighted in bold.}
\label{tab:rouge-bert}
\end{table}

\subsection{Judging Feedback}

\begin{table}[h!]
    \centering
    \begin{tabular}{c|r|r}
        \hline
        \textbf{Criteria} & $\kappa_\text{IAA}$ & $\kappa_\text{AJA}$\\ \hline
        C1   & 0.208   & 0.211\\ 
        C2   & 0.556    & 0.476 \\ 
        C3   & 0.468   & 0.583\\
        C4   & 0.287   & 0.139\\ 
        C5   & 0.395  & 0.236\\\hline
        Overall   & 0.427  & 0.382\\ \hline
    \end{tabular}
    \caption{\textbf{Inter-annotator Agreement Table.} Inter-Annotator Agreement $\kappa_\text{IAA}$ and Adjudicator--Judge agreement $\kappa_\text{AJA}$, both calculated using the Cohen's Kappa statistic. Rows C1--C5 represent agreement for the specified feedback criteria, and row ``Overall'' represents the agreement over all the criteria.}
    \label{tab:feedback-agreement}
\end{table}

To evaluate the quality of the LLM-as-a-judge for feedback evaluation, two of this paper's authors annotated all 48 MAGIC--human feedback pairs for each of the five criteria (C1--C5) with a label of ``LLM'' or ``Human'' to decide a winner for each criterion. A third author then adjudicated any split votes to determine the ``ground truth'' winner, so each criterion winner had at least 2 human votes. We then computed the Cohen's Kappa between the two annotators and between the adjudicator and the judge LLM as shown in Table \ref{tab:feedback-agreement}. The overall (across C1--C5) Inter-Annotator Agreement was moderate ($\kappa_\text{IAA}=0.427$)  while the overall Adjudicator--Judge agreement was fair ($\kappa_\text{AJA}=0.382$).

Our analysis of LLM-as-a-judge shows considerable agreement between human-adjudicated preferences and LLM (o4-mini) preferences for C2 and C3, indicating that the LLM judge can reliably assess which feedback has better analysis of strengths and weaknesses. Meanwhile C1, C4, and C5 had slight to fair agreement. This might be due to these criteria being more subjective and to the feedback quality being roughly similar between human and MAGIC. Finally, while the Judge LLM prefers the MAGIC feedback 69\% of the time, the adjudicator only gave MAGIC preference 52\% of the time, averaged across C1--C5, which could be due to the positive character of the prompt used for our study \cite{koutcheme_evaluating_2024}. 

\section{Conclusion}
We introduced MAGIC, a zero-shot framework for multi-trait AES and AEF. We evaluated MAGIC on a set of collated GRE essays with feedback ground truths and demonstrated the effectiveness of our approach. To the best of our knowledge, the utility of LLMs for both zero-shot AES and AEF in L1 and L2 college-level writing had not been fully demonstrated prior to this work, breaking the long-standing over-reliance on ASAP and variants.

We find that MAGIC's orchestrated multi-agent design improves scoring agreement relative to a single agent baseline. To accomplish this, we developed an enhanced version of the GRE dataset where each of the five traits was scored for every essay. We then assessed both overall agreement with GRE holistic scores and per-agent reliability using annotated per-trait scores, demonstrating consistent performance on both levels. 

Furthermore, we assess MAGIC's feedback quality using different methods. We propose an LLM judge to automatically evaluate the feedback at scale, and then evaluate its agreement with human preferences by manually annotating feedback pairs. We observe an overall fair agreement between judge and human decisions, and conclude that MAGIC generated feedback has quality comparable to humans.

The dataset's high quality enabled us to formally characterize feedback with respect to its source essay. In particular, we developed a metric for feedback specificity, a novel measurement for AEF research. Based on the Jaccard similarity scores between a ground truth essay and its associated feedback, we pose a threshold where generated feedback should outperform a random, yet valid, text, by at least one standard deviation, to affirm that its advice is targeted rather generic. MAGIC exceeds baseline feedback under this measure, but its outputs could be improved with clearer and guided prompting. Finally, we find that MAGIC’s resulting feedback is comparable to human feedback in length, semantic relevance, and lexical diversity measures, further strengthening our claim that the generated feedback is human-aligned.

\section{Future Work}
The availability of a large corpus containing a mixture of L1 and L2 speech, ground truth scores per-trait and feedback remains to be seen, and is of utmost importance for future work in AES and AEF.  Quality feedback and essays are already rare, and essays with feedback for languages outside of English are even more so. The MAGIC framework can be tested on more diverse samples to demonstrate true generalizability. 

The evaluation of these feedback technologies in a real classroom setting is also key for deploying any education tooling, and while our orchestrator mitigates generic feedback, further calibration techniques or human-in-the-loop pipelines could be explored.

Future work on MAGIC would include blinded and expanded evaluation of the feedback quality experiments using additional external graders or comparisons of multiple LLM judges for robustness. We would also explore ablations showing agent contributions and orchestration robustness for different LLM families and parameter sizes.

\section*{Ethical Statement}
Publicly available material of sample essays, scores, and feedback was collated from publicly distributed legacy ETS study material to build the ground truth evaluations. The corpus does not contain personally identifiable or sensitive information.

All the texts and scores collated as ground truth are accessible under Fair Use guidelines and have been made freely available to the public. However, ETS holds formal copyright of GRE material and has not yet approved publication of a standalone data set.

The authors offer this work as a bridge to deepen participation and discussion between students and instructors to motivate the development of critical thinking and writing skills. However, this type of work has the potential to be abused by bad actors to disrupt standardized testing environments by providing unapproved feedback or false scores. 

\section*{Acknowledgments}
The authors appreciate the insight, guidance, and support of Narges Norouzi and Gireeja Ranade, who piloted AI for Education courses at UC Berkeley's Computer Science department for the 2024-2025 academic year. We also thank Nelson Lojo for providing feedback and polish to the drafts. Modal (modal.com) generously supplied research credits to access server infrastructure to run exploratory and test experiments. 

\bibliography{references}

\end{document}